# Classifying Shelf Life Quality of Pineapples by Combining Audio and Visual Features

Yi-Lu Jiang[1], Wen-Chang Chang[2], Ching-Lin Wang[3],
Kung-Liang Hsu[3], Chih-Yi Chiu[3]*

[1]Department of Horticulture, National Chiayi University, No.300 Syuefu Rd.,
Chiayi City, 600355, Taiwan (R.O.C.).

[2]Department of Food Science, National Chiayi University, No.300 Syuefu Rd.,
Chiayi City, 600355, Taiwan (R.O.C.).

[3]Department of Computer Science and Information Engineering,
National Chung Cheng University, No.168, Sec. 1, University Rd.,
Minhsiung, Chiayi, 621301, Taiwan (R.O.C.).

*Corresponding author(s). E-mail(s): cychiu@cs.ccu.edu.tw ;
Contributing authors: jinagyl@mail.ncyu.edu.tw ;
wcchang@mail.ncyu.edu.tw ; asz18351@gmail.com ;
mike90331@gmail.com;

**KEYPOINTS / HIGHLIGHTS**

- A multimodal and multiview dataset (PQC500) of 500 pineapples was constructed using tapping audio and visual images.
- A contrastive audiovisual masked autoencoder (CAV-MAE) was adapted for shelf life classification of pineapples.
- The proposed cross-modal model achieved 84% accuracy, outperforming unimodal models by up to 18%.
- Audio features from side-tapping with unidirectional microphones were found to be the most effective.
- The proposed framework is scalable and can be extended to other crops for quality grading and waste reduction.

**IMPACT**

This study contributes to the field of precision agriculture by introducing an automated, non-destructive classification framework for assessing pineapple shelf life using multimodal data. The proposed method enhances post-harvest decision-making and has the potential to reduce food waste and labor costs across the supply chain.

**KEYWORDS**

contrastive audiovisual and masked autoencoder, multimodal, multiview, shelf life, waste reduction

**ABSTRACT**

Determining the shelf life quality of pineapples using non-destructive methods is a crucial step to reduce waste and increase income. In this paper, a multimodal and multiview classification model was constructed to classify pineapples into four quality levels based on audio and visual characteristics. For research purposes, we compiled and released the PQC500 dataset consisting of 500 pineapples with two modalities: one was tapping pineapples to record sounds by multiple microphones and the other was taking pictures by multiple cameras at different locations, providing multimodal and multi-view audiovisual features. We modified the contrastive audiovisual masked autoencoder to train the cross-modal-based classification model by abundant combinations of audio and visual pairs. In addition, we proposed to sample a compact size of training data for efficient computation. The experiments were evaluated under various data and model configurations, and the results demonstrated that the proposed cross-modal model trained using audio-major sampling can yield 84% accuracy, outperforming the unimodal models of only audio and only visual by 6% and 18%, respectively.

## 1. INTRODUCTION

Pineapple is one of the main edible fruits and economic crops, known for its high nutritional value and rich dietary fiber. After banana and citrus, the pineapple production is the third largest in the world, cultivated in numerous tropical and subtropical countries, such as the Philippines, Thailand, Indonesia, Malaysia, Kenya, India, and China (Ali et al., 2020). The global market for pineapples has seen a significant increase in demand, which presents attractive international business prospects. Pineapple can generally be stored for a couple of weeks at room temperature and will generate a variety of flavors as time passes. From farm to table, it can be consumed as fresh fruit or processed into juice and canned food. Additionally, it serves as a primary source for extracting alcohol and producing livestock goods, especially in the context of utilizing industrial waste.

The evaluation of the quality of pineapple plays a crucial role in influencing consumer preferences, handling after harvest, and determining the fruit's market value. Although advances in agricultural technology have significantly improved the quantity and quality of pineapples, production is greatly affected by various weather conditions. Furthermore, extreme climate effects will complicate production management. A typical scenario is the harvest period in summer. Extremely elevated temperatures and intense rainfall raise the percentage of pineapples with a short shelf life, which are prone to rot and unsuitable for storage and transportation. This could result in a large backlog of unsold pineapples and cause them to be discarded.

To reduce the waste of pineapples, people seek to evaluate the shelf life of pineapples by non-destructive approaches. Once pineapples are harvested from the farm, their qualities are immediately assessed and sorted for appropriate processing and storage: pineapples with a longer shelf life can be transported to fresh produce markets, while those with a shorter shelf life should be immediately consumed or processed in food factories. Conventional assessing approaches include tapping the pineapple with a rubber stick or finger to listen to the sound, or visually inspecting the pineapple's appearance. However, these methods are highly dependent on expert experience. In addition, during the peak season, there will be significant manpower and time to inspect thousands of pineapples every day. Efficiency and accuracy can be compromised as work hours extend. The reduction in manpower demand is very beneficial for labor-intensive tasks.

The growing demand for high-quality pineapples has recently increased research interest in quality assessment. However, *only a few pineapple datasets are publicly available*. The website (Images.cv, n.d.) provided photos and drawings for pineapple and non-pineapple classification. In Kaggle, a pineapple dataset was contributed for object detection provided by Adhil (2022). The dataset contained image data of raw and ripe pineapples with the object detection label format. Kalabarige et al. (2024) augmented the Kaggle pineapple dataset and defined four maturity classes for classification using transfer learning and multihead attention models. These public datasets provided only the visual modality that was mainly used for image classification and object detection tasks; they lacked the necessary information for pineapple quality classification. We released a new pineapple dataset containing rich audio and visual data on GitHub.

Traditional techniques for quality assessment, such as chemical and physical analyses, are often destructive, labor-intensive, and time-consuming. Therefore, people are interested in non-destructive and cost-effective inspection, driving the need for automation and machine learning techniques. Ali et al. (2023) gave a good review paper that presents the study of quality attributes of pineapples and evaluates related technologies like spectroscopy, computer vision, acoustic, and instrument-based sensing. In particular, computer vision-based approaches that used cost-effective devices and time-efficient inspection provide a promising avenue. To identify the translucency disorder, Haff et al. (2006) took X-ray images for expert inspection, while Xu et al. (2022) and Qiu et al. (2023) used visible and near-infrared spectra with machine learning classification. Tantinantrakun et al. (2023) and Semyalo et al. (2024) also employed near-infrared spectroscopy but was used to predict the maturity level of pineapple. Dittakan et al. (2018) extracted local binary pattern of the pineapple skin as a texture descriptor to classify the pineapple grade in terms of sweetness and juiciness. Chang et al. (2022) and Siricharoen et al. (2023) presented an object detection framework to detect pineapples in photos and identify the sweet and sour taste. Among various machine learning techniques, deep learning has attracted increasing attention for fruit grading recently. Chuquimarca et al. (2024) offered an evaluation of convolutional neural networks (CNN) applied to assess external quality in image-based fruit classification for ripeness, deformities, and defects.

The use of acoustic data for quality classification has become popular recently because of its precision and convenience. Huang et al. (2022) defined the pineapple quality as two classes, namely, drum sound and meat sound (same as hollow sound and solid sound defined later respectively). By recording the tapping sound of pineapple, its acoustic characteristic is transformed into the spectrogram and classified by CNN. Phawiakkharakun et al. (2022) and Phawiakkharakun & Pongpinigpinyo (2024) also employed the tapping sound for pineapple quality classification. They defined quality as the degree of juiciness of three levels and adopted CNN and ensemble learning models. To judge the maturity level of pineapple, Pathaveerat et al. (2008) applied multivariate data analysis for destructive chemical parameters and non-destructive acoustic impulse response and fruit weights measured in air and water. Arwatchananukul et al. (2024) evaluated six classifiers in machine learning to analyze acoustic responses, discovering that the random forest algorithm surpassed the other methods. Chen et al. (2022) designed a device that injects excitation sound into pineapple and an acoustic coupler to receive vibration energy, which was then processed by wavelet kernel decomposition and clustering with AdaBoost for ripeness classification.

Note that the *above methods mainly consider one modality for pineapple quality classification*. In particular, the work of Huang et al. (2022) is closely related to ours. The two methods are both centered on predicting shelf life, where Huang et al. (2022) distinguishes only two classes, while we propose a more granular approach by defining four distinct classes. More significantly, *our method leverages the abundance of audio and visual data to explore the possibilities of utilizing multimodal and multiview analysis*.

To minimize manual effort in quality classification, we leveraged machine learning techniques and compile a multimodal and multiview dataset for training. The dataset was termed PQC500 consisting of 500 pineapples with two modalities: one was tapping pineapples to record sounds using multiple microphones and the other was taking pictures by cameras at different locations, providing multimodal and multiview audiovisual features for analysis and as a learning corpus. With the diverse and detailed information of the dataset, we constructed a classification model based on a state-of-the-art representation learning scheme called contrastive audiovisual masked autoencoder (CAV-MAE) proposed by Gong et al. (2022), which fused tapping sounds and appearance images to learn a joint audiovisual representation. Our experiment evaluated various combinations of modalities and views of audio and visual features, and the result validated the effectiveness of the proposed multimodal and multi-view model in achieving the highest accuracy. By integrating the novel machine learning technique into the conventional harvest procedure, pineapples can be processed immediately and distributed appropriately. It thus ensured optimal shelf life preservation and reduces significant labor effort, creating a better revenue model for producers. In addition, the AI-assisted framework can be applied to other crops to achieve the goal of reducing food waste and promoting the circular agriculture economy.

We highlight the main contributions as follows:
- We compiled and published the PQC500 dataset to classify the shelf life quality of pineapples. PQC500 contains 10000 audio soundtracks and 8000 photos of different tapping and views collected from 500 pineapples, each of which is labeled with the horticultural classification levels. This is the first dataset of pineapples with multimodal and multiview features so far. The dataset is available on GitHub: http://github.com/ncyuMARSLab/PQC500.
- We constructed a classification model in a multimodal and multiview learning approach. The model

can be improved by leveraging various combinations of audio and visual pairs of different views. To our knowledge, it is the first model that fuses multiple modalities and views to predict pineapple quality and manifests its high potential and wide applicability to other crops.

- We conduct numerous experiments to evaluate the performance differences under unimodal and multimodal modes. The findings indicate that the audio modality surpasses the visual modality, and the multimodal model exhibits superior performance compared to the unimodal model. In addition, we examine various classification models across different sampling strategies and sample sizes, which yield valuable insights and recommendations.

The remainder of this paper is organized as follows. In Section 2, we present the data acquisition process and method used for non-destructive quality classification of pineapples, especially those based on computer vision and acoustic approaches. Section 3 demonstrates and discusses the experimental results. Section 4 provides interpretation and discusses key observations from the results. Conclusion remarks are given in Section 5.

## 2. MATERIALS AND METHODS

PQC500 was a dataset that comprised audio and visual features of pineapples. Its goal was to contribute towards classifying the shelf life quality of pineapples. In the following subsections, we detailed the specifications for data acquisition and analysis.

### 2.1 Data acquisition

To create the pineapple dataset, we collected 500 pineapples of the Tainung No. 17 variety, which was the most widely produced species in Taiwan. To facilitate data acquisition, we designed a device that was embedded with several sensors. The device had a platform connected to a mechanically driven rubber strip and was surrounded by microphones and cameras. The layout of the device and sensors was shown in Fig. 1. The microphones and cameras were listed on two sides of the platform. Two microphones and one camera were placed at location-1; their facing directions were the same as the tapping direction. Three microphones and one camera were placed at location-2; their facing directions were perpendicular to the tapping direction. The microphones had two different types, namely unidirectional and omnidirectional. The unidirectional microphone mainly recorded sound in front of the microphone and was less susceptible to background noise, whereas the omnidirectional microphone recorded all surrounding sound. In this study, the brands of unidirectional and omnidirectional microphones were RODE VideoMicro (super-cardioid polar pattern) and Logitech webcam, respectively. The microphone sampling rate was 48000 Hz. The camera photo resolution had two sizes: $1280 \times 720$ (at location-1) and $1920 \times 1280$ pixels (at location-2).

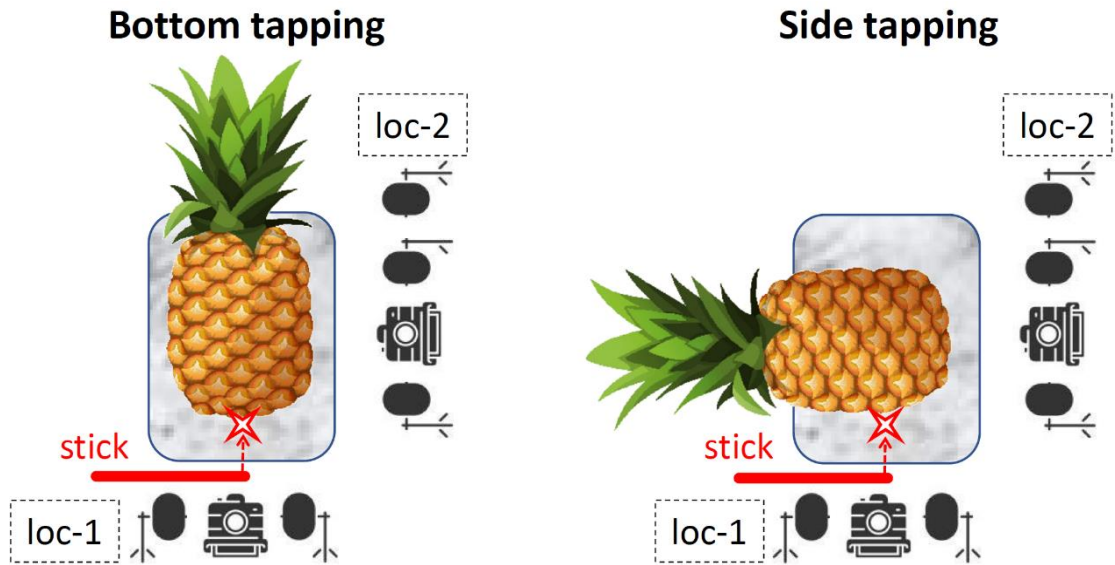

Fig. 1    The layout of the device and sensors that extracted audio and visual features of pineapples for bottom tapping (left) and side tapping (right).

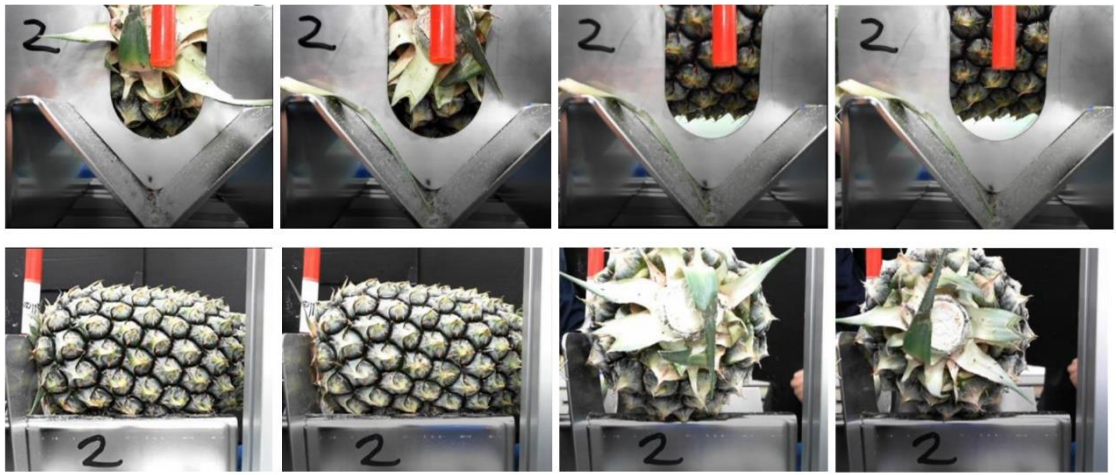

Fig. 2    Example of photos of a pineapple captured at different camera locations and tapping surfaces. The first row photos were captured by the location-1 camera, and the second row photos were captured by the location-2 camera. The first two columns showed bottom tapping photos, and the last two columns showed side tapping photos.

When a pineapple was put on the platform, it was tapped on the side and bottom surfaces, and the sound was recorded by the surrounding microphones and cameras at the two locations. The pineapple was tapped four times, twice for the side surface and twice for the bottom surface. Each time, the pineapple was rotated randomly to augment data variation. Fig. 2 gave an example of photos of a pineapple captured at different camera locations and tapping surfaces. The first and second rows showed the photos captured in location-1 and location-2, respectively. The first two columns showed bottom tapping photos, and the last two columns showed side tapping photos. The duration of the recorded soundtrack was 3 to 10 seconds. Consequently, each pineapple produced 20 soundtracks (recorded with five surrounding microphones) and 16 photos (captured by two cameras at different locations), where soundtracks were stored in the WAV format and photos were saved in the JPEG format. In total, PQC500 contained 10000 audio files and 8000 color image files, which constituted a multimodal and multiview dataset of audio and visual features with different views.

The shelf life quality of pineapples could be simply classified into two categories: drum sound and meat sound, referring to the sound when tapping the pineapple. As the names suggested, the drum sound was heard like beating a drum, which meant that the pineapple contained less water (low moisture) and its flavor was usually sweet-and-sour. The meat sound was heard as beating meat, which meant that the pineapple contained more water (high moisture) with a juicy taste. The classification could be used to anticipate the shelf life roughly: the drum sound pineapple could be stored longer, while the meat sound pineapple was going to rot. However, according to the horticultural area, we employed a fine-grained category set in this study, which provided four grading levels to help people make more precise decisions. Therefore, the following four labels for shelf life quality were defined in this study: (1) hollow sound (abbreviated as H); (2) semi-hollow sound (SH); (3) semi-solid sound (SS); and (4) solid sound (S). The order also represented the moisture content from low to high. In fact, the hollow sound and solid sound were equivalent to the above-mentioned drum sound and meat sound, respectively. However, using horticultural labels could provide fine-grained information to describe the pineapple quality more precisely. To determine the label for each pineapple, we adopted the water selection approach. That was, a pineapple was placed in water to observe its position and angle for labeling, as illustrated in Fig. 3. Consequently, each pineapple was associated with one of the four labels.

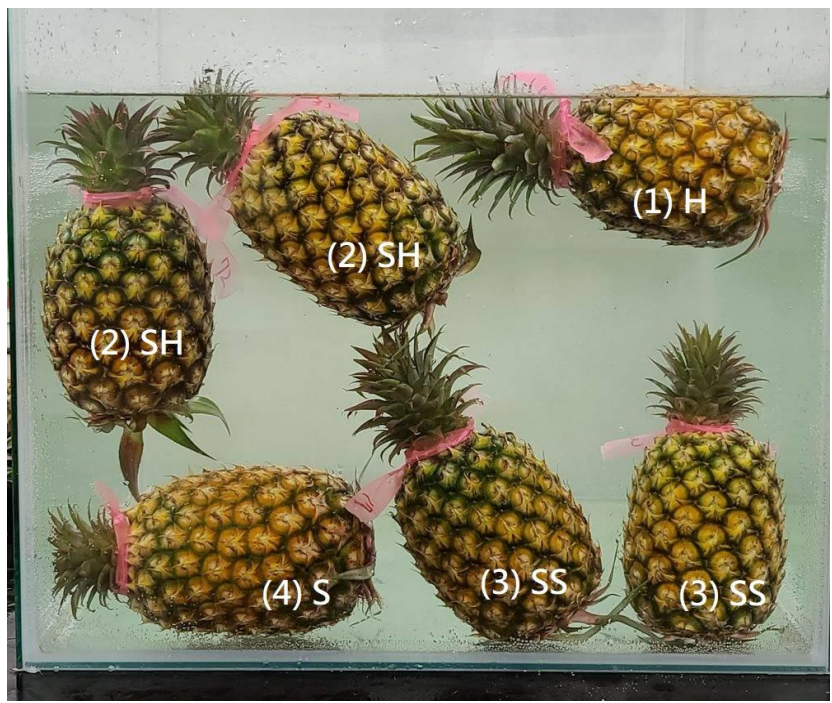

Fig. 3    Example of determination of pineapple quality by the water selection approach.

## 2.2 Data analysis

To make the experimental data consistent, we divided PQC500 into training and test sets in a 4:1 ratio. The numbers of soundtracks and photos in the training and test sets were listed in Table 1. Stratified random sampling was applied to build the training and test sets based on the data distribution among the four quality labels to reflect the population being studied. As shown in Figure 5, it was clear that the distribution was highly imbalanced; some machine learning tricks could be applied to deal with the imbalanced situation.

PQC500 provided multimodal and multiview features for each individual pineapple. In this study, the definitions for multimodal and multiview were given as follows. Multimodal meant multiple modalities of data, such as audio and images. Multiview meant data were collected from multiple sources. For example, several microphones of different types were placed around the pineapple to record tapping sounds at various positions. Taking into account the richness and diversity of multiple modalities and views, we combined their information to perform better on various learning tasks.

Table 1 The number of soundtracks and photos of the training and test data.

|  | #Pineapples | #Soundtracks | #Photos |
|---|---|---|---|
| Training | 400 | 8000 | 6400 |
| Test | 100 | 2000 | 1600 |
| Total | 500 | 10000 | 8000 |

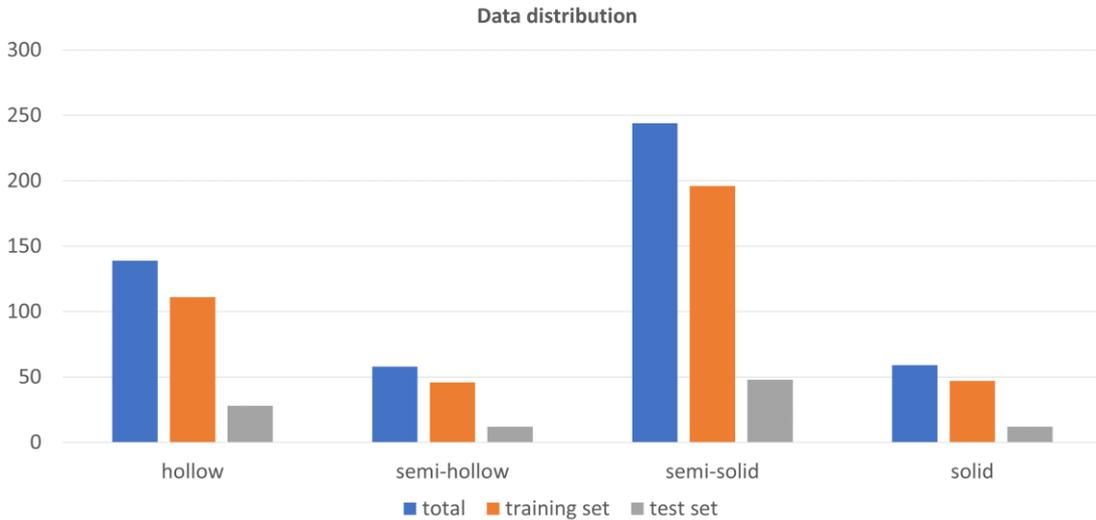

Fig. 4    The data distribution among the four quality labels.

## 2.3 Multimodal and multiview learning

To fully utilize the multimodal and multiview data of PQC500, the prediction model needed to be able to process audio and visual features and learn their mutual correlation efficiently. We considered two multimodal learning frameworks, namely ensemble learning and cross-modal learning. The ensemble learning framework built a heterogeneous ensemble consisting of different models of modalities, where each model was trained independently but concatenated with each other at the last stage. For example, a stacking ensemble combined predictions or representations from the audio model and the visual model. The cross-modal learning framework aggregated information from different modalities at the early stage, so that the modalities could learn from each other to generate a fused representation. For comparison purposes, we implemented some backbone models in this study, including the vanilla CNN and ResNeXt (Xie et al., 2017) models for ensemble learning and the CAV-MAE model (Gong et al., 2022) for cross-modal learning, as detailed below.

Let $T = \{\{(P^{(i)}, L^{(i)})\}\}$ be the training data of PQC500, where $P^{(i)}$ was the i-th training sample and $L^{(i)}$ was the corresponding label, $i \in [1, I]$, and $I$ was the number of training samples. Let $a_j^{(i)}$ and $v_k^{(i)}$ be the j-th audio feature and the k-th visual feature of $P^{(i)}$, where $j \in [1, J]$, $k \in [1, K]$, $J$ and $K$ were the numbers of audio features and visual features, respectively. We took the ResNeXt model as an example to demonstrate the ensemble learning process; the vanilla CNN model could be adopted with the same process. The ResNeXt model was an extension of the ResNet model (He et al., 2016), which was CNN with residual connections between different layers. ResNeXt was designed to process the modality of image data. To process the modality of audio data, we converted time-series audio signals into 2D feature maps and regarded them as specific image data. Therefore, we trained two ResNeXt models: one for processing the audio features and the other for processing the visual features. The outputs of the ResNeXt models were concatenated and followed by a multi-layer perceptron (MLP) block for classification, where the MLP block consisted of two fully connected layers with a ReLU activation function in between. The ResNeXt model for ensemble learning could be expressed as Eq. 1:

$$\begin{aligned}
f_{a_j}^{(i)} &= \text{ResNeXt}(a_j^{(i)}). \\
f_{v_k}^{(i)} &= \text{ResNeXt}(v_k^{(i)}). \\
f_{a_j v_k}^{(i)} &= \text{Concatenate}\left(f_{a_j}^{(i)}, f_{v_k}^{(i)}\right). \\
\Upsilon^{(i)} &= \text{Softmax}(\text{MLP}(f_{a_j v_k}^{(i)})).
\end{aligned} \quad (1)$$

where $f_{a_j}^{(i)}$, $f_{v_k}^{(i)}$ and $f_{a_j v_k}^{(i)}$ were the audio representation, visual representation, and audiovisual joint representation respectively, and $\Upsilon^{(i)}$ was the prediction output of class probabilities for $P^{(i)}$.

It was natural to process multiple modalities using cross-modal analysis models such as Zhu et al. (2022), Gong et al. (2022), and Chen et al. (2024). In this study, we implemented the CAV-MAE model (Gong et al., 2022) as the backbone for our purpose. CAV-MAE integrated contrastive learning and masked data modeling to learn a joint and coordinated audiovisual representation. It followed the encoder-decoder architecture. The encoder concatenated the audio and visual modality streams for contrastive learning, and the decoder masked the joint audiovisual representation of the encoder output for reconstruction learning. The joint audiovisual representation of the encoder output could be used for downstream tasks such as classification and retrieval. To meet our classification task, we concatenated the pre-trained encoder of CAV-MAE with a randomly initialized linear classification head to perform supervised learning. Given the audio feature and the visual feature, the CAV-MAE encoder learned their mutual information and transformed them into a joint audiovisual representation, followed by an MLP block for classification and regression. The modified CAV-MAE classifier for cross-modal joint learning was represented by Eq. 2:

$$f_{a_j v_k}^{(i)} = CAV - MAE(a_j^{(i)}, v_k^{(i)}).$$

$$\Upsilon^{(i)} = Softmax\left(MLP\left(f_{a_j v_k}^{(i)}\right)\right). \quad (2)$$

Figure 5 showed the frameworks of ensemble learning and cross-modal learning.

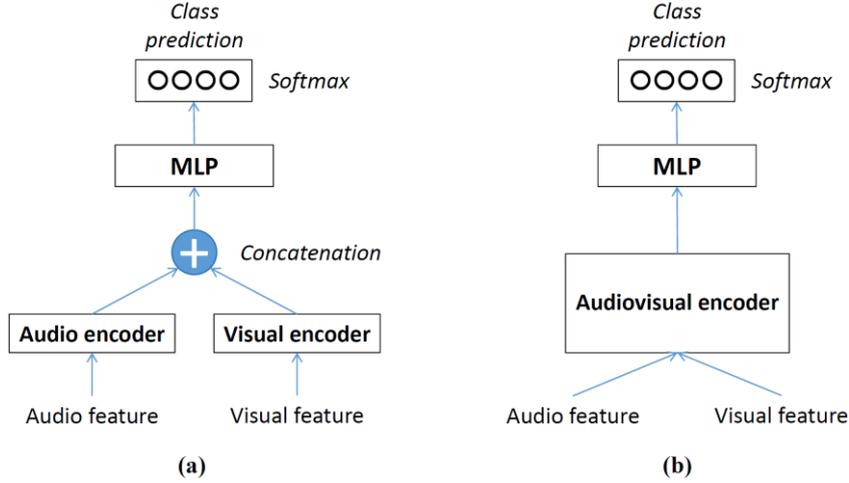

Fig. 5   The proposed frameworks: (a) ensemble learning and (b) cross-modal learning.

## 2.4 Data preprocessing

The audio and image content of PQC500 was acquired and archived in a raw form; it had to be organized into a state in which it could be easily used in the proposed frameworks. Given the raw tapping sound of the audio file, we normalized the amplitude signal to the interval [0, 1] and detected the maximum peak amplitude along the time domain. Based on the maximum peak, we cropped the signal by only keeping the segment from 0.1 seconds before to 0.3 seconds after the peak, which contained clear tapping and echo sounds and removed most of the background noise. The 0.4-second segment was resampled at 22050 Hz and converted to the Mel spectrogram of the 2D feature map $1024 \times 128$, denoted as the audio input feature $f_{a_j}^{(i)}$ used in Eq. 1. To preprocess the pineapple image file, since the camera was capturing from a fixed direction, it was easy to crop the image region to retain the pineapple and automatically remove unrelated background content. The cropped color image was resized to $224 \times 224$ pixels, denoted as the visual input feature $f_{v_k}^{(i)}$.

## 2.5 Training data sampling

Multiview data sources could be used to increase the richness and variation of the data. By combining different modalities and views, plentiful training data pairs could be generated effectively. Recall that in the multimodal learning framework, the input form was a pair of audio and visual features, unlike the unimodal learning framework where only a single-modality feature was used during the training process. Although the natural pairing of audio and visual instances in pineapples provided useful information for learning audiovisual representations, the complexity of multimodal learning was usually higher than that of unimodal learning. In addition to learning complexity, a substantial number of training instance pairs could be generated by permuting the audio and visual instances for multimodal learning. For example, given $J$ audio files and $K$ image files for each of $I$ training samples, in unimodal learning, the number of training instances was $I \times J$ soundtracks for an audio classification model, and $I \times K$ photos for a visual classification model. In multimodal learning, the number of training instance pairs was the product of the instances of the two modalities, resulting in a total of $I \times J \times K$ audiovisual pairs. In the PQC500 training set, we had $I = 400$, $J = 20$, and $K = 16$. In unimodal learning, the audio classification model had $I \times J = 8000$ training instances of soundtracks, and the visual classification model had $I \times K = 6400$ training instances of photos. However, in multimodal learning, we could pair each soundtrack with each photo, thus generating $I \times J \times K = 128000$ audiovisual training pairs. It was clear that there existed an efficiency bottleneck when training entire audiovisual pairs in multimodal learning.

We proposed to sample training instance pairs to improve training efficiency. Intuitively, a high-quality training instance would contribute to better performance. Instead of using all instances for training, selecting a subset of high-quality instances could consume fewer computational resources without sacrificing too much accuracy. To identify high-quality instances, the unimodal learning results offered useful insights. As shown in the experimental section, we had the following observations in the unimodal learning framework:

- For the tapping surface, the side tapping instances performed more accurately than the bottom tapping instances.
- For the audio modality, the location-1 soundtracks (closer to the tapping stick) performed more accurately than the location-2 soundtracks.
- For the visual modality, the bottom surface photos performed more accurately than the side surface photos.

These observations could be used to guide the design of the pineapple tapping device and the configuration of the sensors. Considering a practical situation where we had a limited budget to equip the device with only one microphone and one camera, a heuristic layout would have involved placing the microphone at location-1 to record the side-tapping sound and directing the camera to capture the bottom surface. A similar rationale applied to selecting high-quality training data pairs, which could be formed by combining the location-1 soundtracks and the bottom region photos. We conducted more comprehensive experimental studies to compare different sampling strategies for composing training instance pairs, as elaborated in the subsequent section.

## 3. RESULTS

Our experiments utilized the PQC500 dataset, where multimodal and multiview data sources provided diverse evaluation modes for training and testing. The classification models were implemented using the PyTorch library, and all experiments were run on a personal computer with an Intel Core i7-14700 CPU, 64 GB of RAM, and an NVIDIA 4070 Ti Super display card with 16 GB of GPU memory. To address the imbalanced class distribution and interclass relations in the PQC500 dataset, we applied class weighting and label smoothing techniques in the implementation. This was accomplished by configuring the PyTorch function CrossEntropyLoss, where the weight parameter was set to the reciprocal of the relative size of each class to alleviate the class imbalance, and the label_smoothing parameter was enabled to address interclass relation issues. The experiments were divided into unimodal and multimodal modes, and the corresponding results and discussions were presented in the following sections.

### 3.1 Unimodal mode

In the unimodal mode, a single audio or visual modality was used for training and testing. The training data contained 400 pineapples, each of which had 20 soundtracks and 16 photos, resulting in a total of 8000 soundtracks and 6400 photos used to train the audio and visual modality models, respectively. The test data contained 100 pineapples, with 2000 soundtracks and 1600 photos used to evaluate the corresponding models. Three backbone models were implemented for each modality, including CNN, ResNeXt, and CAV-MAE. The CNN model was based on the implementation of Huang et al. (2022), which consisted of three blocks, each comprising a convolution layer followed by a max pooling layer. Note that Huang et al.'s CNN model was originally designed for the audio modality only. We followed the same architecture to build a corresponding CNN model for the visual modality. For the ResNeXt model, we used ResNeXt-50, which consisted of fifty concatenated residual blocks. For the CAV-MAE model, we used the unimodal version, in which only the audio or visual encoder was employed for training and testing. We first examined the visual modality. The training set was divided according to the camera location (location-1 or location-2) and the photo content (bottom surface or side surface). The corresponding test set was used to assess accuracy, which was measured based on the definition of the confusion matrix. Let *M* be a confusion matrix, where *M(i, j)* represented the number of test instances whose actual classes were *i* and whose predicted classes were *j*. Then the accuracy was calculated based on the summation of the main diagonal expressed as:

$$accuracy = \frac{\sum_{i=1}^{4} M(i,i)}{\sum_{i=1}^{4} \sum_{j=1}^{4} M(i,j)} \qquad (3)$$

Table 2 listed the accuracy of the visual modality. We gave the following observations:

- The bottom surface photos yielded higher accuracy than the side surface photos.
- The location-1 photos showed better performance in the CNN and ResNeXt models, while the location-2 photos performed better in the CAV-MAE model.

- The CAV-MAE model outperformed both the CNN and ResNeXt models.

Table 2 The accuracy of the visual modality in the unimodal mode.

| model | Training set | | Accuracy |
|---|---|---|---|
| | Camera location | Photo content | |
| CNN | Location-1 | Bottom | 0.62 |
| CNN | Location-1 | Side | 0.60 |
| CNN | Location-2 | Bottom | 0.60 |
| CNN | Location-2 | Side | 0.55 |
| ResNeXt | Location-1 | Bottom | 0.64 |
| ResNeXt | Location-1 | Side | 0.55 |
| ResNeXt | Location-2 | Bottom | 0.57 |
| ResNeXt | Location-2 | Side | 0.61 |
| CAV-MAE | Location-1 | Bottom | 0.66 |
| CAV-MAE | Location-1 | Side | 0.65 |
| CAV-MAE | Location-2 | Bottom | 0.64 |
| CAV-MAE | Location-2 | Side | 0.64 |

For the audio modality, the training set was divided according to the tapping surface and microphone location. Table 3 listed the accuracy of the audio modality and revealed these findings:

- The side tapping yielded higher accuracy than the bottom tapping.
- The microphones at location-1 yielded higher accuracy than those at location-2. It was observed that the location indicated the distance from the tapping sound source to the microphone, with location-1 being closer to the source and location-2 being farther away.
- *The CNN model achieved the highest accuracy under the configuration of side tapping with a location-1 microphone, surpassing both the ResNeXt and CAV-MAE models.*

Since the side tapping demonstrated better performance than the bottom tapping, in the next experiments, we fixed the tapping to the side surface and investigated the impact of sound quality. Recall that our data were collected using two types of microphones, namely unidirectional and omnidirectional. To understand which type performed better, we designed the following experiment using different sound qualities, including unidirectional, omnidirectional, and both for model training. The results were listed in Table 4. *The omnidirectional sound quality demonstrated inferior performance compared to the unidirectional sound quality. Using both types of sounds could enhance performance in certain cases, such as in the CNN model, which achieved the best accuracy.*

Table 3 The accuracy of the audio modality in the unimodal mode.

| model | Training set | | Accuracy |
|---|---|---|---|
| | Tapping surface | Microphone location | |
| CNN | Bottom | Location-1 | 0.69 |
| CNN | Bottom | Location-2 | 0.70 |
| CNN | Side | Location-1 | 0.78 |
| CNN | Side | Location-2 | 0.72 |
| ResNeXt | Bottom | Location-1 | 0.65 |
| ResNeXt | Bottom | Location-2 | 0.65 |
| ResNeXt | Side | Location-1 | 0.71 |
| ResNeXt | Side | Location-2 | 0.68 |
| CAV-MAE | Bottom | Location-1 | 0.72 |
| CAV-MAE | Bottom | Location-2 | 0.74 |
| CAV-MAE | Side | Location-1 | 0.72 |
| CAV-MAE | Side | Location-2 | 0.73 |

Table 4 The accuracy of the audio modality with different sound quality in the unimodal mode.

| model | Training set | | | Accuracy |
|---|---|---|---|---|
| | Tapping surface | Microphone location | Sound quality | |
| CNN | Side | Location-1 | Unidirectional | 0.76 |
| CNN | Side | Location-1 | Omnidirectional | 0.55 |
| CNN | Side | Location-1 | Both | 0.78 |
| CNN | Side | Location-2 | Unidirectional | 0.76 |
| CNN | Side | Location-2 | Omnidirectional | 0.58 |
| CNN | Side | Location-2 | Both | 0.72 |
| ResNeXt | Side | Location-1 | Unidirectional | 0.75 |
| ResNeXt | Side | Location-1 | Omnidirectional | 0.48 |
| ResNeXt | Side | Location-1 | Both | 0.71 |
| ResNeXt | Side | Location-2 | Unidirectional | 0.64 |
| ResNeXt | Side | Location-2 | Omnidirectional | 0.48 |
| ResNeXt | Side | Location-2 | Both | 0.68 |
| CAV-MAE | Side | Location-1 | Unidirectional | 0.73 |
| CAV-MAE | Side | Location-1 | Omnidirectional | 0.60 |
| CAV-MAE | Side | Location-1 | Both | 0.72 |
| CAV-MAE | Side | Location-2 | Unidirectional | 0.69 |
| CAV-MAE | Side | Location-2 | Omnidirectional | 0.52 |
| CAV-MAE | Side | Location-2 | Both | 0.73 |

In summary, the audio modality yielded better accuracy than the visual modality. This explained why people typically assessed pineapple quality by employing the tapping method rather than relying on visual inspection. Moreover, domain knowledge could be used to explain the result: the bottom surface photo showed the stem status, which might be highly related to the quality, and the side tapping sound could provide a clearer internal echo than the bottom tapping sound. Another observation revealed that the CNN model achieved the greatest accuracy when employing the side-tapping sound feature in the unimodal mode. The ResNeXt model, nevertheless, suffered from overfitting, delivering accurate predictions for the training data but failing to perform well on the test data. We argued that the ResNeXt model comprised an excessive number of parameters relative to the available data.

## 3.2 Multimodal mode

In the multimodal mode, we evaluated and compared the proposed frameworks of ensemble learning and cross-modal learning. The ensemble learning framework implemented the vanilla CNN and ResNeXt models, and the cross-modal learning framework implemented the CAV-MAE model. As large combinations of training instance pairs could be generated by multiview data sources, we focused on how to sample effective audio and visual features under constrained computing resources. Let $\Omega^{(i)} = \{ a_j^{(i)}, v_k^{(i)} \mid j = 1,2,\ldots,J \text{ and } k = 1,2,\ldots,K \}$ be the set of audio-visual pairs of pineapple $P^{(i)}$, where $i = 1,2,\ldots,I$. The size of $\Omega^{(i)}$, i.e., the number of audio-visual pairs, was $|\Omega^{(i)}| = J \times K$. Instead of using the whole set, we sampled a part of $\Omega^{(i)}$ for cross-modal learning to reduce the training time cost. We denoted the number of sampled pairs from $\Omega^{(i)}$ as $S$, and the total number of sampled pairs across $I$ training pineapples was $I \times S$. In this experiment, we set $I = 400$ and varied $S$ to observe the performance with different numbers of training sample pairs. In addition, the following sampling strategies were considered:

- Random sampling. $S$ samples were randomly selected from $\Omega^{(i)}$.
- Audio-major sampling. $S$ samples were selected from $\Omega^{(i)}$, where particular audio features $a_j^{(i)}$ were sampled and paired with different visual features. The audio-major sampling strategy ensured that each selected audio feature was used at least once during training.
- Visual-major sampling. $S$ samples were selected from $\Omega^{(i)}$, where particular visual features $v_k^{(i)}$ were sampled and paired with different audio features. The visual-major sampling strategy ensured that each selected visual feature was used at least once during training.

Based on the discussion given in Section 2.5, we selected the audio features from the location-1 microphones for audio-major sampling and the visual features from the location-2 camera for visual-major sampling.

The test set was compiled based on similar considerations. To make the pineapple tapping device affordable to most users, its design needed to be compact. Suppose that only one camera and one microphone could be installed. The optimal layout could be arranged by placing the location-1 microphone and the location-2 camera for side tapping, which provided the most effective audio and visual sources, as shown in Figure 6. We simulated this scenario to prepare the test instances by pairing the location-1 soundtracks recording the side tapping sound with the location-2 photos capturing the bottom surface. Note that there were two microphones at location-1, each of which recorded two soundtracks, whereas the camera at location-2 captured four photos. Pairing four soundtracks with four photos generated sixteen audio-visual pairs for each pineapple, resulting in a total of 1600 audio-visual pairs compiled from 100 test pineapples as the test set.

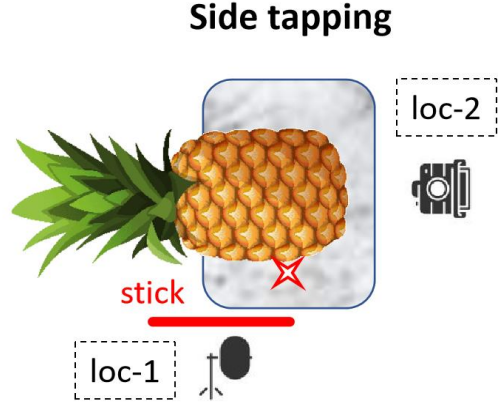

Fig. 6 The proposed optimal layout of the device and sensors for the test environment.

Table 5 listed the comparison for the CNN, ResNeXt, and CAV-MAE models, each of which adopted different sampling strategies of random, audio-major, and visual-major. In addition, we varied the number of training samples to observe the relationship with accuracy. Key findings were as follows.

From the perspective of the learning models, the CAV-MAE model yielded the best performance among the three models. The accuracy reached 0.84, which marked a notable improvement over the 0.74 obtained from the audio modality in the unimodal mode. CAV-MAE was designed for cross-modal analysis of audio and visual features, making it naturally suitable for learning mutual information from audiovisual pairs. However, CAV-MAE was a complex model that employed a Transformer-based architecture and was trained using contrastive learning and masking techniques. More training data were required to learn optimal model weights. Hence, in the multimodal mode, by providing sufficient training instance pairs, the performance of CAV-MAE was effectively enhanced. Another reason for the improvement was that the unimodal CAV-MAE retained only one branch either audio or visual whereas the multimodal CAV-MAE incorporated both modalities, allowing each to serve as a soft label for the other and thus providing richer information than one-hot labels. On the other hand, the CNN and ResNeXt models showed smaller improvements in accuracy from the combination of the two modalities. These models were originally designed to process single-modality image data. To handle both audio and visual modalities simultaneously, the two models were modified via ensemble learning by stacking the audio and visual branches in parallel and concatenating their final representations for classification. Since the fusion of the two modalities was performed only at the final stage, mutual influence was not involved in the intermediate layers, thereby weakening the effectiveness of cross-modal learning. Nevertheless, CNN performed better than ResNeXt due to the overfitting issue in ResNeXt.

Table 5 The accuracy of training instance sampling in the multimodal mode.

| model | Training set | | Accuracy |
|---|---|---|---|
| | Sampling strategy | #Samples( I×S) | |
| CNN | Random | 1600 | 0.75 |
| CNN | Random | 3200 | 0.79 |
| CNN | Random | 6400 | 0.74 |
| CNN | Random | 12800 | 0.76 |
| CNN | Audio-major | 1600 | 0.75 |
| CNN | Audio-major | 3200 | 0.76 |
| CNN | Audio-major | 6400 | 0.75 |
| CNN | Audio-major | 12800 | 0.74 |
| CNN | Visual-major | 1600 | 0.72 |
| CNN | Visual-major | 3200 | 0.77 |
| CNN | Visual-major | 6400 | 0.69 |
| CNN | Visual-major | 12800 | 0.68 |
| ResNeXt | Random | 1600 | 0.75 |
| ResNeXt | Random | 3200 | 0.78 |
| ResNeXt | Random | 6400 | 0.72 |
| ResNeXt | Random | 12800 | 0.50 |
| ResNeXt | Audio-major | 1600 | 0.73 |
| ResNeXt | Audio-major | 3200 | 0.73 |
| ResNeXt | Audio-major | 6400 | 0.71 |
| ResNeXt | Audio-major | 12800 | 0.55 |
| ResNeXt | Visual-major | 1600 | 0.69 |
| ResNeXt | Visual-major | 3200 | 0.69 |
| ResNeXt | Visual-major | 6400 | 0.68 |
| ResNeXt | Visual-major | 12800 | 0.51 |
| CAV-MAE | Random | 1600 | 0.77 |
| CAV-MAE | Random | 3200 | 0.76 |
| CAV-MAE | Random | 6400 | 0.77 |
| CAV-MAE | Random | 12800 | 0.73 |
| CAV-MAE | Audio-major | 1600 | 0.81 |
| CAV-MAE | Audio-major | 3200 | 0.84 |
| CAV-MAE | Audio-major | 6400 | 0.83 |
| CAV-MAE | Audio-major | 12800 | 0.78 |
| CAV-MAE | Visual-major | 1600 | 0.76 |
| CAV-MAE | Visual-major | 3200 | 0.79 |
| CAV-MAE | Visual-major | 6400 | 0.78 |
| CAV-MAE | Visual-major | 12800 | 0.75 |

Next, we considered the performance of different sampling strategies. In general, the audio-major strategy yielded the highest accuracy, while the visual-major strategy outperformed the random strategy. The random strategy served as a baseline for comparison with the other strategies. For example, in the case of the CAV-MAE model, the audio-major and visual-major strategies improved accuracy by approximately 7% and 2%, respectively, over the random strategy. As previously discussed, in the set of audio-visual pairs $\Omega^{(i)}$, the quality of each pair varied. The target-specific sampling methods prioritized certain target instances to generate a high-quality training dataset, whereas the random sampling method treated all instances equally, resulting in an averaged training data quality. In particular, the audio-major strategy, which ensured that all audio features were sampled, increased both the quality and diversity of the training data, thereby enhancing model generalization and accuracy. Since the audio feature demonstrated its superiority over the visual feature in the previous experiments, we suggested that, given a fixed sample size, the audio feature should be prioritized for sampling during training.

Finally, we varied the number of training samples by setting $S = \{4, 8, 16, 32\}$, corresponding to $I \times S = \{1600, 3200, 6400, 12800\}$ training samples. The results showed that accuracy generally peaked at 3200 samples. This indicated that using more training samples beyond that point was unnecessary; selecting an appropriate number of training samples could not only achieve high accuracy but also reduce training time. Notably, the ResNeXt model deteriorated significantly at 12800 training samples. We considered that the overfitting issue in ResNeXt intensified as the number of training samples increased. In conclusion, a small number of high-quality training samples outperformed a large number of medium-quality samples, indicating that sample quality had a greater influence on model accuracy than sample quantity.

## 4. DISCUSSION

We take some examples to analyze the performance on shelf life quality classification. Fig. 7 shows confusion matrices of classification accuracy and t-SNE plots of data visualization for three examples: the left is ResNeXt of the audio modality (accuracy: 0.68); the middle is CAV-MAE of the audio modality (accuracy: 0.72); and the right is CAV-MAE of the audio and visual modalities (accuracy: 0.81). ResNeXt reveals that the predictions are predominantly focused on the two primary categories of hollow sound and semi-solid sound. This occurs due to the tangled and scrambled data distribution of the semi-hollow and solid sound classes, making them challenging to distinguish from other categories. CAV-MAE gives better predictions for the four classes, and the data distribution can be approximated by a straight line. Indeed, according to our quality definition, the four labels can be seen as representing different levels of moisture content, which generally align along a linear progression.

In our implementation, we only enabled the parameters in the CrossEntropyLoss function to address the imbalanced class distribution and interclass relation issues. In fact, we tried other techniques such as data augmentation (using the torchvision library), class-balanced loss (proposed by Cui et al. (2019)), and adaptive label smoothing (proposed by Zhou et al. (2023)). These methods, nonetheless, did not result in a noticeable enhancement of the classification models. This remains a potential area for further study.

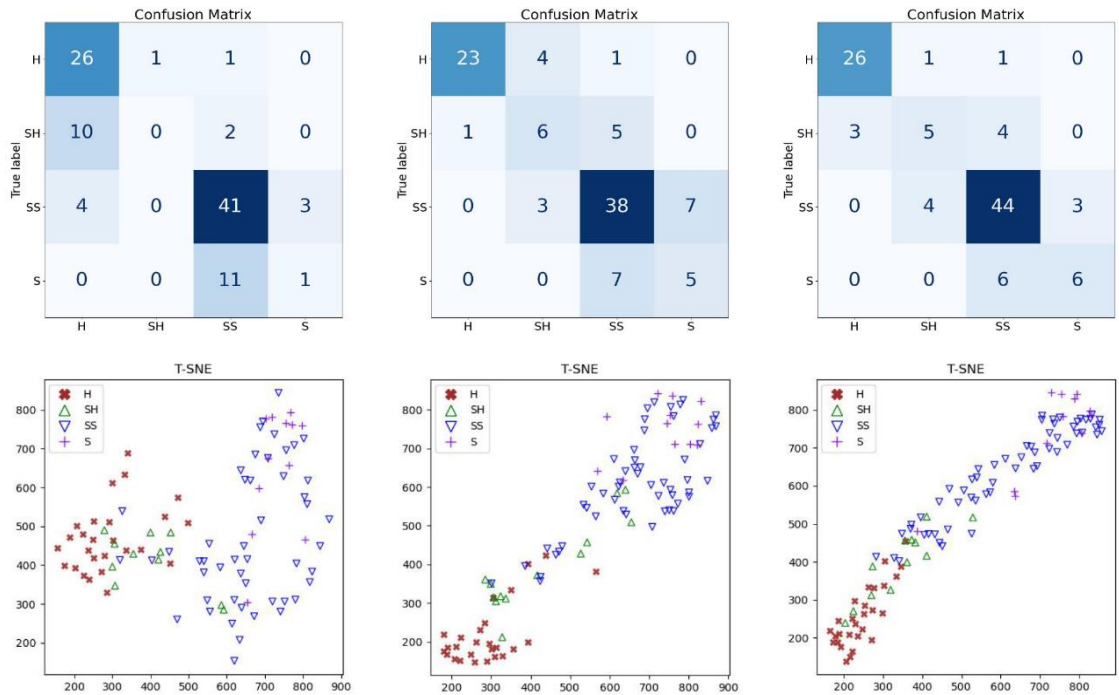

Fig. 7 Examples of confusion matrices and t-SNE plots (best viewed in color). Left: ResNeXt of the audio modality. Middle: CAV-MAE of the audio modality. Right: CAV-MAE of the audio and visual modalities.

## 5. CONCLUSION

This paper introduces a framework that employs multimodal and multiview learning techniques to assess the shelf life quality of pineapples. This novel framework leverages the recently released PQC500 dataset, which includes audio and visual data gathered from multiple sensors at different locations. Comprehensive experiments reveal that the cross-modal model, trained with diverse and abundant audio-visual pair combinations, can achieve outstanding performance.

## 6. CONFLICT OF INTEREST

This work was supported by the National Science and Technology Council under grants NSTC 112-2221-E-415-008-MY3, NSTC 113-2634-F-194-001, and Ministry of Agriculture under grants 112-2.2.1-2.3-001-006.